\documentclass[conference, a4paper]{IEEEtran}

\IEEEoverridecommandlockouts
\usepackage{graphicx}
\usepackage[english]{babel}
\usepackage[utf8]{inputenc} 
\usepackage[T1]{fontenc}
\usepackage{amssymb}
\usepackage{booktabs}    
\usepackage[table]{xcolor} 
\usepackage{tikz}
\usepackage{cite}

\ifCLASSINFOpdf
  \usepackage{graphicx}
\else
\fi

\usepackage[cmex10]{amsmath}
\usepackage{multirow}
\usepackage{array}
\usepackage[lofdepth,lotdepth]{subfig}
\usepackage{xcolor}
\usepackage{makecell}
\usepackage{rotating}
\usepackage{threeparttable}
\usepackage{colortbl} 
\usepackage{rotating} 
\usepackage{multirow} 
\usepackage{adjustbox} 
\usepackage{tikz} 
\usepackage{array} 
\usepackage{float} 
\usepackage{afterpage}
\usepackage{placeins}
\usepackage{xurl} 
\usepackage{hyperref}

\AtBeginDocument{%
  \renewcommand\tablename{TABLE}
}

\AtBeginDocument{%
  \renewcommand\abstractname{Abstract}
}

\usepackage{calc}

\begin{document}

\title{
WeatherOcc3D: VLM-Assisted Adverse Weather Aware 3D Semantic Occupancy Prediction}

\author{
    \IEEEauthorblockN{A. Enes Doruk}
    \IEEEauthorblockA{Department of Artificial Intelligence and Data Engineering \\
    Ozyegin University \\
    Istanbul, Türkiye \\
    enes.doruk@ozu.edu.tr}
    \and
    \IEEEauthorblockN{Abdelaziz Hussein}
    \IEEEauthorblockA{Department of Artificial Intelligence and Data Engineering \\
    Ozyegin University \\
    Istanbul, Türkiye \\
    abdelaziz.hussein@ozu.edu.tr
    }
    \and
    \IEEEauthorblockN{Hasan F. Ates}
    \IEEEauthorblockA{Department of Artificial Intelligence and Data Engineering \\
    Ozyegin University \\
    Istanbul, Türkiye \\
    hasan.ates@ozyegin.edu.tr}
}
\maketitle

\begin{abstract}
While multi-modal 3D semantic occupancy prediction typically enhances robustness by fusing camera and LiDAR inputs, its effectiveness is fundamentally constrained by environmental variability. Specifically, camera sensors suffer from severe low-light degradation, while LiDAR sensors encounter significant backscatter noise during heavy precipitation. These adverse conditions create a modality trust problem, as static fusion strategies fail to adaptively re-weight inputs when a specific sensor becomes unreliable. To address this, we propose a VLM-assisted framework leveraging the pre-trained CLIP latent space to guide multi-sensor integration via linguistic environmental cues. We utilize a parameter-efficient adapter to align weather-specific text embeddings with sensor features, coupled with a gating strategy that decomposes environmental uncertainty into two factors: visibility and illumination. This enables the model to dynamically modulate the fusion ratio—prioritizing semantic camera features in clear daylight and shifting to geometric LiDAR priors during rainy nights. Evaluations on the nuScenes dataset demonstrate the versatility of our approach, as implementing our proposed framework on the OccMamba and M-CONet architectures achieves mIoU scores of 26.3 and 21.1, respectively, significantly outperforming their traditional baselines.

\end{abstract}
\begin{IEEEkeywords}
Semantic occupancy prediction, Vision language models, Autonomous driving
\end{IEEEkeywords}

\IEEEpeerreviewmaketitle

\IEEEpubidadjcol

\section{INTRODUCTION}
3D semantic occupancy prediction is a cornerstone of autonomous driving, providing a volumetric representation essential for safe navigation. To achieve robustness, modern systems increasingly rely on multi-modal frameworks integrating camera and LiDAR data. While classic techniques such as BEVFusion \cite{liu2023bevfusion} and OccFormer \cite{zhang2024occformer} establish strong baselines through static feature concatenation, their performance is fundamentally constrained by environmental variability. In real-world scenarios, cameras suffer from extreme visibility degradation in heavy precipitation, while LiDAR sensors encounter significant backscatter noise.

These adverse conditions create a modality trust problem. Current state-of-the-art strategies, including attention-based models like OccFusion \cite{zheng2024occfusion} and GaussianOcc3D \cite{doruk2026gaussianocc3d}, attempt to resolve this through dynamic weighting. However, these methods often incur prohibitive memory costs and typically treat sensor inputs as equally reliable regardless of atmospheric noise. Furthermore, while recent VLM-based trends like LanguageOcc \cite{chen2025languageocc} and VEON \cite{zheng2024veon} utilize vision-language embeddings for semantic priors, they rarely address the fundamental problem of sensor trust in degraded conditions.

Unlike previous cross-attention methods that are memory-intensive, our framework introduces a VLM-guided factorized gating mechanism. We utilize a text-prompting strategy to encode environmental descriptors—such as rainy night or clear day—into a shared latent space. These linguistic cues are then decomposed into independent visibility and illumination factors to dynamically modulate the fusion ratio between camera and LiDAR voxels. This fine-grained control allows the model to selectively suppress noise-contaminated channels while preserving high-resolution spatial information, significantly reducing false-positive occupancy predictions caused by sensor artifacts. 
Evaluations on the nuScenes-OpenOccupancy \cite{wang2023openoccupancy}  dataset demonstrate that our approach effectively mitigates sensor-specific degradation across multiple architectures. By integrating our method into two distinct baselines, M-CONet \cite{wang2023openoccupancy} and OccMamba \cite{li2025occmamba}, we achieve improved mIoU scores of 21.1 and 26.3, respectively. Specifically, our method shows superior performance in the most challenging adverse weather subsets, outperforming traditional fusion baselines and establishing a more robust paradigm for multi-modal perception.

\begin{figure*}[ht]
	\centering
	\includegraphics[scale=0.45]{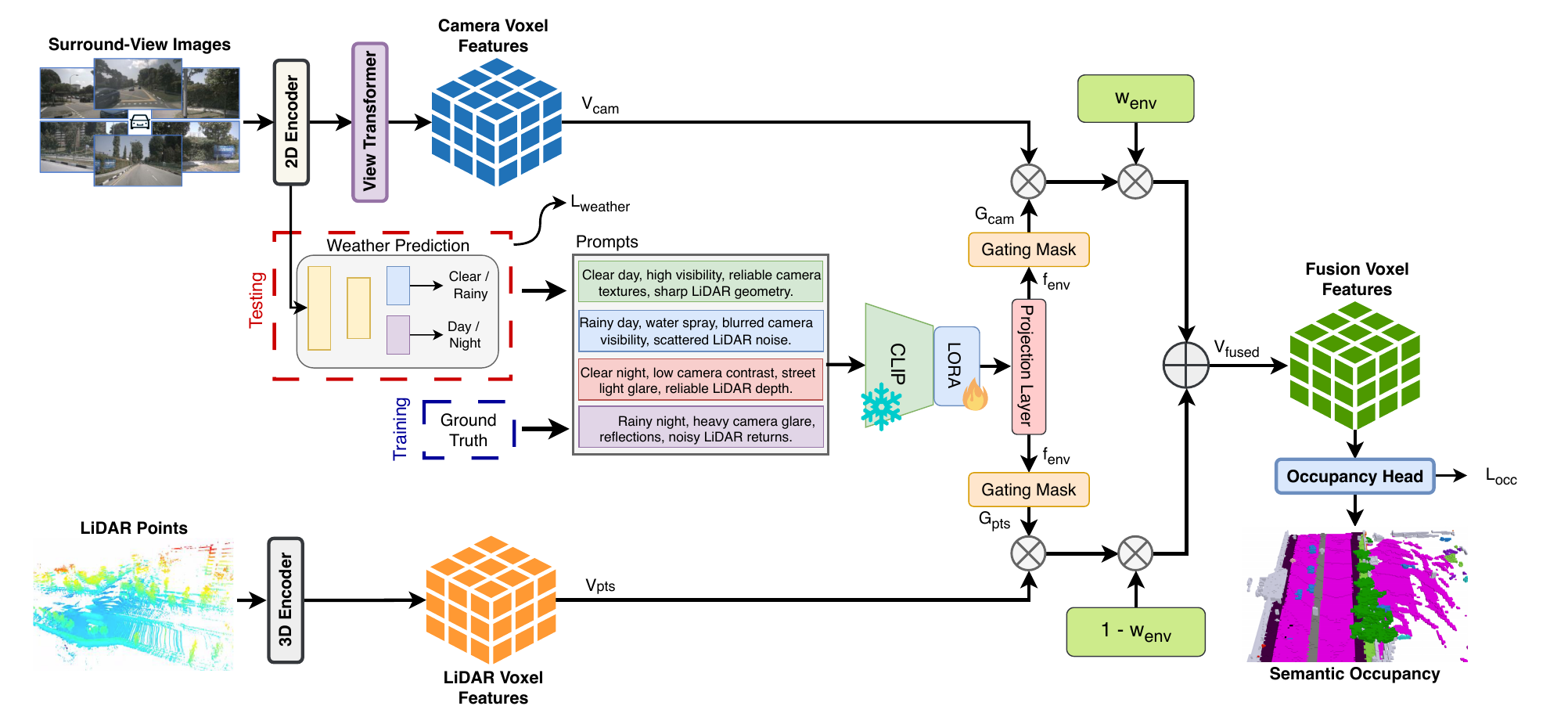}
	\caption{Overview of our proposed model architecture.}
	\label{fig:overview}
\end{figure*}

\vspace{-0.3cm}
\section{RELATED WORK}
The evolution of 3D semantic occupancy prediction has transitioned from single-modality baselines to sophisticated multi-modal fusion strategies.  Classic fusion techniques, such as BEVFusion \cite{liu2023bevfusion} and OccFormer \cite{zhang2024occformer}, establish strong baselines via static concatenation or summation but fail to adapt when environmental factors render a modality unreliable. Modern fusion approaches seek dynamic adaptability through attention-based architectures. Frameworks like OccFusion \cite{zheng2024occfusion}, GaussianOcc3D \cite{doruk2026gaussianocc3d} utilize spatial cross-attention to align features, yet suffer from high computational costs and memory overhead on dense grids, often requiring heavy downsampling that sacrifices fine-grained geometric detail. Recent VLM-based trends, including LanguageOcc \cite{chen2025languageocc}, and VEON \cite{zheng2024veon}  leverage vision-language embeddings to provide rich semantic priors; however, these methods primarily focus on open-vocabulary recognition rather than addressing the fundamental modality trust problem in adverse weather. Although VLMFusionOcc3D \cite{doruk2026vlmfusionocc3d} employs a CLIP-based weather-aware fusion, it relies on external CAN bus signals for weather data, and this causes it to be static and sensor-dependent. Our work bridges this gap by utilizing a VLM-guided factorized gating mechanism. By decomposing environmental context into independent visibility and illumination factors, we provide a robust, memory-efficient alternative to attention-based fusion that maintains high-resolution occupancy performance in degraded conditions.

\definecolor{barriercolor}{RGB}{255, 0, 0}
\definecolor{bicyclecolor}{RGB}{0, 255, 0}
\definecolor{buscolor}{RGB}{255, 255, 0}
\definecolor{carcolor}{RGB}{0, 0, 255}
\definecolor{construction_vehiclecolor}{RGB}{255, 165, 0}
\definecolor{motorcyclecolor}{RGB}{128, 0, 128}
\definecolor{pedestriancolor}{RGB}{255, 192, 203}
\definecolor{traffic_conecolor}{RGB}{255, 69, 0}
\definecolor{trailercolor}{RGB}{192, 192, 192}
\definecolor{truckcolor}{RGB}{139 ,69 ,19}
\definecolor{driveable_surfacecolor}{RGB}{135 ,206 ,235}
\definecolor{other_flatcolor}{RGB}{160 ,82 ,45}
\definecolor{sidewalkcolor}{RGB}{211 ,211 ,211}
\definecolor{terraincolor}{RGB}{139 ,105 ,20}
\definecolor{manmadecolor}{RGB}{112 ,128 ,144}
\definecolor{vegetationcolor}{RGB}{34 ,139 ,34}

\newcommand{\makecolorcell}[2]{%
    \makecell[b]{\begin{turn}{90}\colorbox{#1}{\hspace{3pt}\rule{0pt}{3pt}} #2\end{turn}}}

\begin{table*}[t]
\centering
\caption{Quantitative comparisons on the nuScenes-OpenOccupancy validation set. C, D, L denote camera, depth and LiDAR, respectively. The best and second-best are in bold and underlined, respectively.}
\vspace{-0.3cm}
\resizebox{0.75\linewidth}{!}{
\setlength{\tabcolsep}{1mm}
\begin{threeparttable}
\begin{tabular}{l|c|c|cccccccccccccccc}
\toprule

\makecell[b]{Method} & \makecell[b]{Input\\Modality}  & \makecell[b]{mIoU} & 
\makecolorcell{barriercolor}{barrier} & 
\makecolorcell{bicyclecolor}{bicycle} & 
\makecolorcell{buscolor}{bus} & 
\makecolorcell{carcolor}{car} & 
\makecolorcell{construction_vehiclecolor}{const. veh.} & 
\makecolorcell{motorcyclecolor}{motorcycle} & 
\makecolorcell{pedestriancolor}{pedestrian} & 
\makecolorcell{traffic_conecolor}{traffic cone} & 
\makecolorcell{trailercolor}{trailer} & 
\makecolorcell{truckcolor}{truck} & 
\makecolorcell{driveable_surfacecolor}{drive surf.} & 
\makecolorcell{other_flatcolor}{other\_flat} & 
\makecolorcell{sidewalkcolor}{sidewalk} & 
\makecolorcell{terraincolor}{terrain} & 
\makecolorcell{manmadecolor}{manmade} & 
\makecolorcell{vegetationcolor}{vegetation} \\

\midrule

TPVFormer~\cite{yuan2023tpvformer} & C  & 7.8 & 9.3 & 4.1 & 11.3 & 10.1 & 5.2 & 4.3 & 5.9 & 5.3 & 6.8 & 6.5 & 13.6 & 9.0 & 8.3 & 8.0 & 9.2 & 8.2 \\
SparseOcc~\cite{tang2024sparseocc} & C  & 14.1 & 16.1 & 9.3 & 15.1 & 18.6 & 7.3 & 9.4 & 11.2 & 9.4 & 7.2 & 13.0 & 31.8 & 21.7 & 20.7 & 18.8 & 6.1 & 10.6 \\
3DSketch~\cite{mitani20003d} & C\&D  & 10.7 & 12.0 & 5.1 & 10.7 & 12.4 & 6.5 & 4.0 & 5.0 & 6.3 & 8.0 & 7.2 & 21.8 & 14.8 & 13.0 & 11.8 & 12.0 & 21.2 \\
AICNet~\cite{li2020anisotropic} & C\&D  & 10.6 & 11.5 & 4.0 & 11.8 & 12.3 & 5.1 & 3.8 & 6.2 & 6.0 & 8.2 & 7.5 & 24.1 & 13.0 & 12.8 & 11.5 & 11.6 & 20.2 \\
LMSCNet~\cite{roldao2020lmscnet} & L  & 11.5 & 12.4 & 4.2 & 12.8 & 12.1 & 6.2 & 4.7 & 6.2 & 6.3 & 8.8 & 7.2 & 24.2 & 12.3 & 16.6 & 14.1 & 13.9 & 22.2 \\
JS3C-Net~\cite{yan2021sparse} & L  & 12.5 & 14.2 & 3.4 & 13.6 & 12.0 & 7.2 & 4.3 & 7.3 & 6.8 & 9.2 & 9.1 & 27.9 & 15.3 & 14.9 & 16.2 & 14.0 & 24.9 \\
Co-Occ~\cite{coocc2024} & C\&L  & 21.9 & 26.5 & 16.8 & 22.3 & 27.0 & 10.1 & 20.9 & 20.7 & 14.5 & 16.4 & 21.6 & 36.9 & 23.5 & 5.5 & 23.7 & 20.5 & 23.5 \\

\midrule

M-CONet \cite{wang2023openoccupancy} & C\&L  & 20.1 & 23.3 & 13.3 & 21.2 & 24.3 & 15.3 & 15.9 & 18.0 & 13.3 & 15.3 & 20.7 & 33.2 & 21.0 & 22.5 & 21.5 & 19.6 & 19.8 \\
M-CONet \textbf{+ Ours} & C\&L  & 21.1 & 24.3 & 14.3 & 22.2 & 25.3 & 16.3 & 16.9 & 19.0 & 14.3 & 16.3 & 21.7 & 34.2 & 22.0 & 23.5 & 22.5 & 20.6 & 20.8 \\
\midrule
OccMamba \cite{li2025occmamba} & C\&L  & \underline{25.2} & \underline{29.1} & \underline{19.1} & \underline{25.5} & \underline{28.5} & \underline{18.1} & \underline{24.7} & \underline{23.4} & \underline{19.8} & \underline{19.3} & \underline{24.5} & \underline{37.0} & \underline{25.4} & \underline{25.4} & \underline{25.4} & \underline{28.1} & \underline{24.9} \\
OccMamba \textbf{+ Ours} & C\&L  & \textbf{26.3} & \textbf{30.2} & \textbf{20.2} & \textbf{26.6} & \textbf{29.6} & \textbf{19.2} & \textbf{25.8} & \textbf{24.5} & \textbf{20.9} & \textbf{20.4} & \textbf{25.6} & \textbf{38.1} & \textbf{26.5} & \textbf{26.5} & \textbf{26.5} & \textbf{29.2} & \textbf{26.0} \\
\bottomrule
\end{tabular}
\end{threeparttable}
}
\label{table_openoccupancy}
\end{table*}

\begin{figure*}[ht]
	\centering
	\includegraphics[scale=0.36]{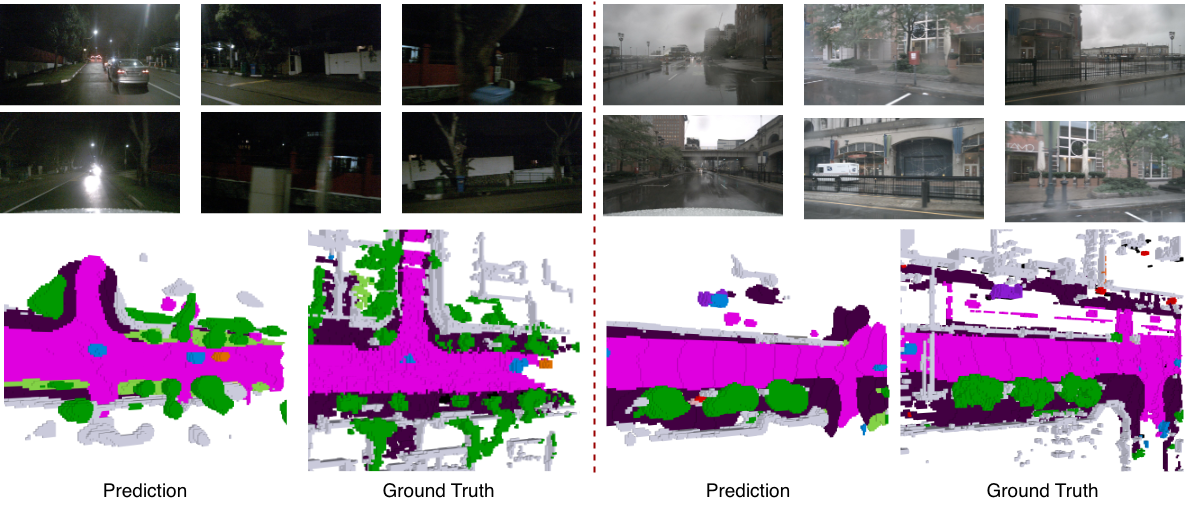}
    \vspace{-0.25cm}
	\caption{Qualitative results under adverse weather and lighting conditions using OccMamba baseline model on the nuScenes-OpenOccupancy validation set.}
	\label{fig:qual}
\end{figure*}

\vspace{-0.3cm}
\section{METHOD}

Our model is an end-to-end multi-modal framework designed for robust 3D semantic occupancy prediction by adaptively fusing camera and LiDAR voxel features through environmental awareness. The model architecture, illustrated in Figure \ref{fig:overview}, begins with the extraction of modality-specific features. Surround-view images are processed through a 2D encoder utilizing a ResNet backbone and a Feature Pyramid Network (FPN)  to extract multi-scale visual features. These 2D features are then projected into a unified 3D space via a View Transformer based on the Lift-Splat-Shoot (LSS) \cite{lss} approach for volumetric reconstruction, forming the camera voxel features $V_{cam} \in \mathbb{R}^{C \times H \times W \times D}$. Concurrently, LiDAR point clouds are processed through a 3D Encoder to generate LiDAR voxel features $V_{pts}$. To drive the adaptive fusion, a weather prediction module analyzes the 2D features using two linear layers that form specialized classification heads. These heads predict environmental states—Clear/Rainy for visibility and Day/Night for illumination—which are supervised during training using ground truth environmental labels. This predictive mechanism is essential to automatically generate the appropriate text prompts during the testing phase.

To translate these descriptors into high-level semantic features, we employ a pre-trained CLIP \cite{clip} text encoder. Because CLIP is trained on a massive, general-purpose dataset, we incorporate Low-Rank Adaptation (LoRA) \cite{lora} to specialize its embeddings for the autonomous driving domain without the prohibitive computational cost of full fine-tuning. We keep the primary CLIP weights frozen and only train the LoRA layers to obtain the refined environment embedding, denoted as $f_{env}$. This embedding $f_{env}$ represents the semantic distillation of the scene's atmospheric conditions. Since the CLIP output does not inherently match our feature space, a Projection Layer consisting of a single linear layer is utilized to map $f_{env}$ into the correct channel dimension. This projected embedding is then passed through a linear layer to generate channel-wise Gating Masks ($G_{cam}, G_{pts}$) that are applied via element-wise multiplication to selectively suppress noise-contaminated channels in each modality.

Following the channel-level gating, the environmental context is used to derive a global trust factor to balance the two modalities. We transform the projected text features into a single learnable scalar $w_{env}$ through a one-layer MLP:
\begin{equation}
w_{env} = \sigma(\text{MLP}(f_{env}))
\end{equation}
where $\sigma$ denotes the sigmoid activation function. This scalar serves as a weather-aware weighting value that represents the model's overall confidence in the visual modality. While the gating masks filter specific noise at the channel level, $w_{env}$ provides a global adjustment to prioritize geometric LiDAR priors when the environmental embedding indicates high atmospheric interference. The final camera trust ratio $w_{env}$ is then used to perform a weighted addition of the gated features. The final fused representation is computed as:
\begin{equation}
V_{fused} = w_{env}  (G_{cam} V_{cam}) + (1 - w_{env})  (G_{pts}  V_{pts})
\end{equation}
\vspace{-0.6cm}

We employ channel-specific feature recalibration, allowing the gating masks to selectively suppress noise-contaminated channels based on the environmental context. Concurrently, the learnable scalar $w_{env}$ serves as a global weighting factor to balance the overall contribution of each modality. While the gating masks perform fine-grained suppression of specific feature dimensions, $w_{env}$ provides a macro-level adjustment to prioritize geometric LiDAR priors when visual features are significantly compromised. The entire framework is optimized through a multi-task objective function:
\begin{equation}
L_{total} = \lambda_{occ} L_{occ} + \lambda_{weather} L_{weather}
\end{equation}
where $L_{occ}$ combines cross-entropy and Lovász-Softmax losses for occupancy prediction, and $L_{weather}$ is the sum of the binary cross-entropy losses from the two environment prediction heads—one for weather (clear or rainy) and the other for time of day (day or night) ensuring accurate prompt selection. This joint supervision ensures that our model dynamically shifts its reliance toward reliable sensor data under adverse conditions.

\section{EXPERIMENTS}
\subsection{Dataset and Implementation}
Our model is evaluated on the nuScenes-OpenOccupancy  dataset, which is based on the nuScenes \cite{caesar2020nuscenes} benchmark. The dataset consists of large-scale multi-modal data collected from 6 surrounding cameras, 1 top LiDAR, and 5 radars, respectively. The occupancy scope is defined from -40m to 40m for the X and Y axes, and -1m to 5.4m for the Z axis. Given a voxel size of 0.4m, the resulting occupancy block shape is $[200, 200, 16]$. For the 3D occupancy prediction task, the official evaluation metric uses the mean Intersection over Union (mIoU) to assess performance:
\begin{equation}
mIoU = \frac{1}{N} \sum_{i=1}^{N} \frac{TP_i}{TP_i + FP_i + FN_i}
\end{equation}
where $TP_i$, $FP_i$, and $FN_i$ represent the number of voxels predicted as true-positive, false-positive, and false-negative for class $i$, respectively, and $N$ is the total number of semantic classes.

We implement our model using the PyTorch framework and conduct training on 4 NVIDIA RTX 6000 GPUs. The model is trained for 20 epochs with a total batch size of 1. For the input modalities, we utilize 10-sweep LiDAR point clouds and multi-view camera images at a resolution of $1600 \times 900$. We employ the AdamW optimizer with an initial learning rate of $2 \times 10^{-4}$. The pre-trained CLIP text encoder remains frozen throughout the training process.

\subsection{Results and Analysis}

\textbf{Comparison with SOTA Methods.}
Table~\ref{table_openoccupancy} presents a comprehensive quantitative comparison of our proposed method against SOTA 3D semantic occupancy prediction models on the nuScenes-OpenOccupancy validation set. Our approach, when integrated with the OccMamba~\cite{li2025occmamba} baseline, achieves a new SOTA performance of 26.3 mIoU, yielding a solid 1.1 mIoU improvement over the vanilla OccMamba architecture. Notably, this performance gain is consistent across all 17 semantic classes. Furthermore, to demonstrate the plug-and-play versatility of our module, we integrated it with M-CONet~\cite{wang2023openoccupancy}, resulting in a similar 1.0 mIoU improvement. This indicates that our fusion strategy generalizes well across different foundational architectures.

\textbf{Analysis of the Fusion Strategy.}
To validate the efficiency of our multi-modal integration, we compare our method against standard and recent fusion techniques in Table~\ref{table:fusion}. Simple operations like Addition and Concatenation yield suboptimal performance (24.9 and 25.2 mIoU). Compared to the recent GaussianOcc3D's ACLF~\cite{doruk2026gaussianocc3d} method, our approach not only achieves a superior segmentation score (+0.8 mIoU) but also operates with a significantly lower latency overhead. Our module adds only 2.14 ms of latency, compared to ACLF's 3.21 ms, demonstrating a highly optimal trade-off between predictive accuracy and computational efficiency suitable for real-time autonomous driving applications.

\vspace{-0.25cm}
\begin{table}[htbp]
\centering
\caption{Performance analysis of different fusion techniques on nuScenes-OpenOccupancy  validation set  with  OccMamba baseline.}
\vspace{-0.3cm}
\resizebox{0.6\linewidth}{!}{ 
\begin{tabular}{l|c|c}
\toprule
Method & mIoU & Latency (ms)\\ 
\midrule
 Addition     & 24.9 & 0.02 \\ 
 Concatenation  & 25.2 & 0.03 \\ 
 ACLF \cite{doruk2026gaussianocc3d}   & 25.5 & 3.21 \\
 \rowcolor{black!10}
Ours    & 26.3 & 2.14 \\
\bottomrule
\end{tabular}
}
\label{table:fusion}
\end{table}
\vspace{-0.2cm}

\textbf{Robustness Under Adverse Weather.}
A critical challenge in multi-modal occupancy prediction is maintaining reliability when environmental conditions degrade sensor fidelity. Table~\ref{tab:weather} details the performance of our weather-aware fusion strategy under varying illumination and precipitation conditions. While the baseline OccMamba~\cite{li2025occmamba} suffers severe performance drops during Night and Rainy scenarios, our method exhibits remarkable resilience. We observe substantial improvements of +3.9 mIoU at night (11.8 to 15.7) and +3.2 mIoU in rainy conditions (24.1 to 27.3). These gains highlight the success of dynamically adapting the fusion weights when the camera stream is compromised by low light or rain artifacts.

\vspace{-0.3cm}
\begin{table}[htbp]
    \centering
    \caption{Performance comparison on the nuScenes-OpenOccupancy validation set under adverse conditions.}
    \vspace{-0.25cm}
    \resizebox{\linewidth}{!}{
     \begin{tabular}{l|c|ccc|ccc}
        \toprule
        & & \multicolumn{3}{c|}{IoU $\uparrow$} & \multicolumn{3}{c}{mIoU $\uparrow$} \\
        Method & Modality  & Rainy & Day & Night & Rainy & Day & Night \\ 
        \midrule
        OccMamba \cite{li2025occmamba} &  C\&L & 28.3 & 36.8 & 12.6 & 24.1 & 26.3 & 11.8 \\
        \rowcolor{black!10}
        OccMamba  \cite{li2025occmamba} \textbf{+ Ours} &  C\&L & 30.2 & 38.1 & 16.2 & 27.3 & 27.1 & 15.7 \\
        \bottomrule
      \end{tabular}
    }
    \label{tab:weather}
\end{table}
\vspace{-0.3cm}

\section{CONCLUSION}
In this paper, we introduced a VLM-assisted framework to resolve the modality trust problem in 3D semantic occupancy prediction under adverse weather. By leveraging a pre-trained CLIP encoder and a factorized gating mechanism, our model dynamically balances camera and LiDAR inputs based on environmental visibility and illumination. Evaluations on the nuScenes-OpenOccupancy dataset demonstrate that our plug-and-play module significantly improves baselines like OccMamba and M-CONet with minimal latency overhead. Ultimately, this adaptive fusion strategy provides a robust, computationally efficient solution for real-world autonomous driving.

{\small
\bibliographystyle{ieeetr}
\bibliography{ref}}

\end{document}